\documentclass[conference]{IEEEtran}

\usepackage{algorithm}
\usepackage{algorithmic}

\usepackage{longtable}

\usepackage{subfigure}
\usepackage{url}
\usepackage{acronym}
\usepackage{subfig}
\usepackage{booktabs}
\usepackage{siunitx}
\usepackage{graphics}
\usepackage{graphicx}
\usepackage{tikz}
\usetikzlibrary{calc}
\usetikzlibrary{shadows}
\usetikzlibrary{shapes}
\usepackage{subfigure}
\usepackage{amsmath}
\usepackage{url}

\usepackage{epstopdf}
\def\R{\mbox{\protect\makebox[.15em][l]{I}R}}

\newcommand{\bi}{\begin{itemize}}\newcommand{\ei}{\end{itemize}}
\newcommand{\be}{\begin{enumerate}}\newcommand{\ee}{\end{enumerate}}
\newcommand{\bs}{\begin{slide}}\newcommand{\es}{\end{slide}}
\newcommand{\bc}{\begin{center}}\newcommand{\ec}{\end{center}}
\newcommand{\beq}{\begin{equation}}\newcommand{\eeq}{\end{equation}}
\newcommand{\beqn}{\begin{eqnarray}}\newcommand{\eeqn}{\end{eqnarray}}
\newcommand{\btab}{\begin{tabular}}\newcommand{\etab}{\end{tabular}}

\ifCLASSINFOpdf
\else
\fi
\begin{document}
%
\title{\vspace*{-0.75cm}{\Large Accepted as a Conference Paper for IJCNN 2016}\\
	Mahalanobis Distance Metric Learning Algorithm for Instance-based Data Stream Classification}


\author{\IEEEauthorblockN{Jorge Luis Rivero Perez}
\IEEEauthorblockA{Department of Informatics Engineering\\Faculty of Science and Technology\\
University of Coimbra\\
Portugal\\
Email: jlrivero@dei.uc.pt}
\and
\IEEEauthorblockN{Bernardete  Ribeiro}
\IEEEauthorblockA{Department of Informatics Engineering\\CISUC-University of Coimbra\\
Portugal\\
Email: bribeiro@dei.uc.pt}
\and
\IEEEauthorblockN{Carlos Morell Perez}
\IEEEauthorblockA{Department of Computer Sciences\\
Faculty of Computer Sciences\\
Central University of Las Villas\\
Cuba\\
Email: cmorellp@uclv.edu.cu}}


%


\maketitle

\begin{abstract}

With the massive data challenges nowadays and the rapid growing of technology, stream mining has recently received considerable attention. To address the large number of scenarios in which this phenomenon manifests itself suitable tools are required in various research fields. Instance-based data stream algorithms generally employ the Euclidean distance for the classification task underlying this problem. 
A novel way to look into this issue is to take advantage of a more flexible metric due to the increased requirements imposed
by the data stream scenario.
In this paper we present a new algorithm that learns a Mahalanobis metric using similarity and dissimilarity constraints in an online manner. This approach hybridizes a Mahalanobis distance metric learning algorithm and a k-NN data stream classification algorithm with concept drift detection. First, some basic aspects of Mahalanobis distance metric learning are described taking into account key properties as well as online distance metric learning algorithms.  Second,  we implement specific evaluation methodologies and comparative metrics such as \emph{Q} statistic for data stream classification algorithms. Finally, our algorithm is evaluated on different datasets by comparing its results with one of the best instance-based data stream classification algorithm of the state of the art. The results demonstrate that our proposal is better in some scenarios and has shown to be competitive in others.
\end{abstract}



%
\IEEEpeerreviewmaketitle

\section{Introduction}
In recent years there has been a great development in the information technology and communications, which has changed the data collection and processing methods \cite{Gama2012,ds_Classif_alg1}. This phenomenon, coupled with the fact that the traditional batch learning has limitations to deal with issues of data stream environments, leads to other data processing techniques. One approach is the instance-based data stream classification algorithms.
Under this scheme, each new instance is compared with existing ones using a distance function, and the closest existing instances are used to assign the class to the new one. However the performance of these methods depends on the quality of the distance function. It is necessary that the function is able to identify the instances that are semantically similar. Likewise, it should also identify as dissimilar those that are semantically different \cite{bellet2013survey}. The general-purpose function does not take into account any statistical regularities that might be estimated from a large training set of labeled examples. However, the best results are obtained when the metric is designed specifically for the task at hand, issue that has received much interest from researchers in the last decade \cite{bellet2013survey,kulis2012metric}. Distance metric learning consists in adapting some pairwise real-valued metric function such as Mahalanobis to the problem of interest using side information as supervision, brought by training examples. Most of methods learn the metric in a supervised manner from similarity, dissimilarity and/or relative distance constraints, being formulated as an optimization problem \cite{bellet2013survey}. Metric learning algorithms have key properties. Each algorithm has properties that define their applicability and suitability for the application at hand. These properties are: Learning paradigm, form of metric, scalability, optimality of the solution and dimensionality reduction. 
Some studies \cite{2knn,3knn,weinberger2006distance} have shown that good design metric learning can significantly improve the k-NN classification accuracy in batch learning. This, with the scalability property of the online distance metric learning, has motivated us to implement a new instance-based data stream classification algorithm, learning a Mahalanobis distance metric. One solution could be to learn the metric in an online way. This approach leads to complex convex optimization problems so it is unable to address well the computational resources restrictions of data stream environments. For that reason we choose KISS Metric Learning algorithm \cite{KISS}, a simple statistical proposal of distance metric learning. We implement a KISSME-based variant (\emph{Keep It Simple and Straightforward MEtric}) in an online setting for hybridizing it with a k-NN algorithm, being our Online-KISSME-Stream the main contribution of this paper. Then, to evaluate its performance we use streaming evaluation methodologies and implement some well comparison metrics for online learning, taking into account aspects such as concept drift detection.
The rest of the paper is organized as follows. In Section~\ref{sec:related} we briefly review the background and related work on Mahalanobis metric learning. In Section~\ref{sec:proposal} we describe and present the above Online-KISSME-Stream. In Section~\ref{sec:experimental} we report and discuss the results of our experiments on three synthetic and one standard real data sets. Finally in Section~\ref{sec:conclusions} the conclusions and outline of future work are discussed.

 

\section{Related Work}
\label{sec:related}
The essence of metric learning is that given a distance function $ d(x_i ,x_j )$ between the data points $x_i, x_j$ lying in some feature space $ X \subseteq \R^d $, for example, the Euclidean one, together with side information as supervision, it should learn a mapping function such that the original distance function applied to the mapped data is better. 
The methods under this approach are dubbed global since they learn a mapping function which is applied to the whole data. Depending on the transformation, this approach is divided into two subclasses: linear and non-linear. For the linear case, the aim is to learn a linear mapping transformation from side information, which may encode the matrix $G$, so that the distance learned is $||Gx_i  - Gx_j||^2$. 
This approach is called Mahalanobis metric learning. The Mahalanobis distance originally refers to a distance measure that incorporates the correlation among the features. However, in the literature this term is used to refer to Mahalanobis generalized quadratic distances defined as: $d_M (x_i ,x_j ) = \sqrt {(x_i  - x_j )^T M(x_i  - x_j )}$ and parameterized by the matrix $ M $ that belongs to the cone of symmetric positive semi-definite (PSD)  $d \times d$  real-valued matrices; as $ M $ is positive semi-definite it can be factorized as $ G^T G $. Then, Mahalanobis distance corresponds to a generalized Euclidean distance using the inverse of the variance-covariance matrix \cite{kulis2012metric,rivero2015}. A general regularized model that captures most of the metric learning existing techniques is proposed in \cite{kulis2012metric}. To encode supervision the author assumes a collection of $ m $ loss functions, which denotes as $c_1,...,c_m$. The other part of the model is a regularizer $ r(M) $ which is a function of $M$. Joining the supervision encoded as loss functions and the regularizer the following model is obtained as a linear combination of these two components: 
\beq
\gamma (M) = r(M) + \lambda \sum\limits_{i = 1}^m {c_i } (X^T MX)
\label{eq:model}
\eeq
where $\lambda$ is a trade-off between the regularizer and the loss, and the goal is  to find the minimum of $\gamma (M)$ over the domain of $ M $, which is the space of PSD matrices. Then the metric learning model is specified as a constrained optimization problem. Quite a few online metric learning algorithms have been proposed, under an approach that learns a Mahalanobis matrix \cite{POLA, LEGO, MDML, RDML}. 
In the sequel we describe some of the most well-known.

\subsection{Online Mahalanobis Metric Learning Algorithms}
POLA (Pseudo-metric Online Learning Algorithm) \cite{POLA}, was the first algorithm focused on online Mahalanobis metric learning that learns a matrix $M$ and a threshold $ b \ge 1 $. POLA in each step $ t $ receives a triplet: $ (x_i, x_j, y_{ij}) $, where $ y_{ij} = 1 $ (if $ (x_i ,x_j ) \in S $ and $ y_{ij} = - 1 $) if $ (x_i ,x_j ) \in D $. The loss function used by POLA is $ c_i (X^T MX) = [1 + y_i (d_M (x_i ,x_j ) - \gamma )]_ +   $ and it uses a squared Frobenius norm for regularization. POLA performs two successive orthogonal projections \cite{kulis2012metric, bellet2013survey}. Another algorithm based on POLA is LEGO (Lego Exact Gradient Online) \cite{LEGO} but it is based on LogDet divergence regularization, which gives it a better performance than POLA. Another algorithm based on POLA is RDML which is more flexible because at each step $ t $ it performs a gradient descent step assuming Frobenius regularization: 
\beq
M^t  = \Pi _{C_ + ^d } (M^{t - 1}  - \lambda y_{ij} (x_i  - x_j )(x_i  - x_j )^T )
\label{eq:two}
\eeq
where $ \Pi _{C_ + ^d } (M^{t - 1}  - \lambda y_{ij} (x_i  - x_j )(x_i  - x_j )^T ) $ is the projection to the PSD cone. The $ \lambda$ parameter implements a trade-off between satisfying the pairwise constraints and keeping close to the matrix of the previous step $ M^{t - 1} $. Unlike POLA, the authors perform the update solving a convex quadratic program \cite{bellet2013survey}. In MDML (Mirror Descent Metric Learning) \cite{MDML} the authors proposed a general framework for online Mahalanobis distance learning. It is based on composite mirror descent and is focused on regularization with the nuclear norm \cite{bellet2013survey}. These algorithms for the best of our knowledge have not been hybridized with a data stream classification algorithm such as k-NN, hence the novelty of our proposal lies in the hybridization and evaluation on data stream environments assuming all the restrictions imposed by this context.
In the majority of these works the authors focus on finding the best combination of regularizers and loss function according the model described by the eq.~\ref{eq:model}. The proposals vary depending on their regularizer and loss function. All these ones receive the supervision information step by step taking into account the dimension time. 
\section{Proposed Approach}
\label{sec:proposal}
In our proposal, we implemented KISS Metric Learning algorithm \cite{KISS} while in an online setting and hybridized it with a k-NN data stream classification algorithm to compute the distance in each query. We compare its performance with IBLStream \cite{2012iblstreams}, which is one of the best instance-based data stream classification algorithms. 
\subsection{KISSME Algorithm}
The KISSME algorithm \cite{KISS} is a simple proposal from a statistical inference point of view, making some assumptions about the distributions to obtain a Mahalanobis matrix from similar and dissimilar constraints. The authors consider two independent generation processes for observed commonalities of similar and dissimilar pairs.

They define the dissimilarity by the plausibility of belonging either to one or the other. Then, from a statistical inference point of view the optimal statistical decision whether a pair $(i,j)$ is dissimilar or not is obtained by a likelihood ratio test. Thus, the authors test the hypothesis $H_0$ that a pair is dissimilar and on the other hand the alternative $H_1$: \cite{KISS}

\beq
\delta (x_i ,x_j ) = \log \left( {{{p(x_i ,x_j |H_0 )} \over {p(x_i ,x_j |H_1 )}}} \right)
\label{eq:three}
 \eeq 

Then, $H_0$ is validated with a high value of $\delta (x_i ,x_j )$. In contrast, a low value means that $H_0$ is rejected and the pair is considered as similar. The authors cast the problem in the space of pairwise differences $(x_{ij}  = x_i  - x_j)$ with zero mean and re-write eq.~\ref{eq:three} to: \cite{KISS}

\beq
\delta (x_{ij} ) = \log \left( {{{p(x_{ij} |H_0 )} \over {p(x_{ij} |H_1 )}}} \right) = \log \left( {{{f(x_{ij} |\theta _0 )} \over {f(x_{ij} |\theta _1 )}}} \right)
\label{eq:four}
 \eeq 

Where $f(x_{ij} |\theta _1 )$ is a probability density function with parameters $\theta _1$ for hypothesis $H_1$ that a pair $(i,j)$ is similar $(y_{ij} = 1)$ and vice-versa $H_0$ for a pair being dissimilar. The authors assume a Gaussian structure of the difference space and relax the problem, obtaining a Mahalanobis distance metric that reflects the properties of the log-likelihood ratio test by re-writing eq.~\ref{eq:four} to: \cite{KISS}

\beq
  \delta \left( {x_{ij} } \right) = \log \left( {\frac{{\frac{1}{{\sqrt {2\pi |\Sigma y_{ij}  = 0|} }}\exp \left( { - {\raise0.7ex\hbox{$1$} \!\mathord{\left/  {\vphantom {1 2}}\right.\kern-\nulldelimiterspace} \!\lower0.7ex\hbox{$2$}}x_{ij}^T \Sigma _{y_{ij}  = 0}^{ - 1} x_{ij} } \right)}}{{\frac{1}{{\sqrt {2\pi |\Sigma y_{ij}  = 1|} }}\exp \left( { - {\raise0.7ex\hbox{$1$} \!\mathord{\left/  {\vphantom {1 2}}\right.\kern-\nulldelimiterspace} \!\lower0.7ex\hbox{$2$}}x_{ij}^T \Sigma _{y_{ij}  = 1}^{ - 1} x_{ij} } \right)}}} \right) 
 \eeq 
where 
\beq \sum\nolimits_{y_{ij}  = 0}^{}  =  \sum\limits_{y_{ij = 0} } {(x_i  - x_j )(x_i  - x_j )^T } 
\label{eq:six}
\eeq
\beq  \sum\nolimits_{y_{ij}  = 1}^{}  =  \sum\limits_{y_{ij = 1} } {(x_i  - x_j )(x_i  - x_j )^T } 
\label{eq:seven}
\eeq
are the similar and dissimilar constraints computed by the outer vector product. Then, by re-projection of its differences: \beq \hat M = \left( {\sum\nolimits_{y_{ij}  = 1}^{ - 1}  -  \sum\nolimits_{y_{ij}  = 0}^1 {} } \right) 
\label{eq:eight}
\eeq
onto the cone of positive semidefinite matrices. Hence, to obtain the Mahalanobis matrix $ M $  they clip the spectrum of $ \hat M $  by eigenanalysis.
\subsection{Online-KISSME-Stream}
For our online variant\footnote{Available in: \url{http://eden.dei.uc.pt/~jlrivero/Online-KISSME-Stream.tar.gz}}, we initialize the k-NN classification algorithm with an Euclidean distance function to compute the distances among the instances setting a diagonal $d \times d$ Mahalanobis matrix, where $d$ is the number of attributes. Then we define the maximun number of instances that can be stored in the base. While the instances base does not store the maximun number each new arriving instance is added to it (algorithm~\ref{alg:algoritmoRaro}, line: 3). The class of this instance is compared with the classes of instances stored previously in the base (algorithm~\ref{alg:algoritmoRaro}, line: 5, 8). If the classes are the same then it is a similarity constraint needed to update the similar matrix calculated by the outer vector product (eq.~\ref{eq:six}) (algorithm~\ref{alg:algoritmoRaro}, lines: 5-7). But if the classes are not the same then it is a dissimilarity constraint and we update the dissimilarity matrix (eq.~\ref{eq:seven}) (algorithm~\ref{alg:algoritmoRaro}, lines: 8-10). When the base stores this number of instances we compute the Mahalanobis matrix by means of the difference between the inverse similarity and dissimilarity matrices (eq.~\ref{eq:eight}) (algorithm~\ref{alg:algoritmoRaro}, lines: 14-15) and we set that the algorithm has learned (algorithm~\ref{alg:algoritmoRaro}, line: 16).  In the next step, we substitute the previous matrix by setting the Mahalanobis matrix computed before. 
After this stage, each arriving instance is classified by the k-NN algorithm with the Mahalanobis matrix learned (algorithm~\ref{alg:algoritmoRaro}, lines: 19-41). The concept drift detection is performed by means of Drift Detection Method \cite{DDM}. Different concept drift levels are evaluated (algorithm~\ref{alg:algoritmoRaro}, lines: 24-30). When a warning level is detected the Mahalanobis matrix is updated again (eq.~\ref{eq:eight}) (algorithm~\ref{alg:algoritmoRaro}, lines: 25-27). But when the level of concept drift is  out control then all the parameters are reset to its defaults values except the Mahalanobis matrix (algorithm~\ref{alg:algoritmoRaro}, lines: 28-30). It means to delete all the instances in the base and also set the algorithm in learning mode again. However the matrix used is the Mahalanobis one previously learned.
Finally, we edit the instances base deleting the instances that have the same label as the arriving instance. The arriving instance is always added.  The pseudo-code of the proposed method is listed in the algorithm~\ref{alg:algoritmoRaro}.
\begin{algorithm}
\begin{algorithmic}[1] 
\REQUIRE Instance, maxbaseSize 

\IF {learned = false} 
 
 \IF {instanceBase $ \prec $  maxbaseSize} 
  \STATE instanceBase.add(Instance)
     
 \FOR{instance in instanceBase}
   \IF {instance.class=Instance.class} 
     \STATE update similarMatrix 

      \STATE similarSize=similarSize+1 
    \ELSE
      \STATE update dissimilarMatrix 

     \STATE dissimilarSize=dissimilarSize+1
   \ENDIF 
 \ENDFOR
\ENDIF  

\IF {instanceBase=maxbaseSize} 
\STATE update mahalanobisMatrix
\STATE learned = true
\ENDIF

\ELSE
\STATE neighbours=search.KNN(Instance)
\STATE makeDistribution (neighbours,distances) 
trueClass=Instance.class
\IF {makeDistribution.maxIndex=trueClass}
\STATE prediction=true
\ENDIF

\STATE update Concept Drift level 
\IF {ddmLevel=ddmWarningLevel}
\STATE update mahalanobisMatrix
\ENDIF
\IF {ddmLevel=ddmOutcontrolLevel}
\STATE resetLearning 
\ENDIF
\IF {prediction=true}
\FOR{neighbourInstance in neighbour}
\IF {Instance.class=neighbourInstance.class}
\STATE deleteInstance()
\ENDIF
\ENDFOR
\ENDIF

\STATE insertInstance(Instance)

\WHILE{learner.size $ \prec $ maxbaseSize} 
\STATE delete the oldest instance from the instanceBase by deleteInstance()
\ENDWHILE
\ENDIF 
\end{algorithmic} \caption{Online-KISSME-Stream}\label{alg:algoritmoRaro} \end{algorithm}

In the next section we present the experimental results obtained from comparing our proposed approach with IBLStream \cite{2012iblstreams}. 
Our study  takes into account the indicators of temporal relevance, spatial relevance and consistency. Subsequent tasks of  editing the database instances to remove or add new instances lead to optimize the composition and size of the case autonomously base. Unlike our proposal, IBLStream instead of learning a distance metric, uses Value Difference Metric (VDM) as distance measure to determine the set of k-NN instances to classify. IBLStream is implemented in MOA (Massive Online Analysis) \cite{Bifet2010} which is an extensible framework, that apart from allowing to implement algorithms also permits  running experiments for online learning.
\begin{table*}
	\centering
	\caption{Experimental data sets characterization.}
	\begin{tabular}{lcccccr}
	\hline
		Dataset & Numeric  Attributes & Nominal Attributes  & Concept drift & Noise &  Number \\		 
	\hline
		Random Tree Generator & 5 & 5 & No & No & 2 \\
		Waveform & 21 & 0 & No & Yes & 3 \\
		SEA & 3 & 0 & No & Yes & 2 \\ 
		Rotating Hyperplane & 10 & 0 & Yes & Yes & 2 \\	
		Random RBF & 10 & 0 & Yes & Yes & 2 \\	
		KDD Cup 99 & 11 & 2 & Yes & Yes & 2 \\	
	\hline
	\end{tabular}
	\label{table:Table1}
\end{table*}
\section{Experimental Evaluations}
\label{sec:experimental}
In the case of streaming classification algorithms, there have been some proposals \cite{Gama2013,Gama2009,shaker2013recovery,bifet2013pitfalls} on evaluation methodologies. In particular, there have been some works on what metrics are appropriate to evaluate the performance of the classifiers. There are basically two data stream evaluation methodologies known as: (i) holdout (ii) prequential. Both of them with forgetting mechanisms. Regarding the latter, sliding windows and fading factors are popular whenever fast and efficient change detection are required. In \cite{Gama2009,Gama2013} it is argued that prequential error with forgetting mechanisms should be used to provide reliable error estimators. Therein it is  proved that, the use of prequential error with forgetting mechanisms reveals to be advantageous in assessing performance and comparing stream learning algorithms. In the design of our experiments we use the prequential evaluation methodology with fading factors as forgetting mechanism and we compute some metrics such as: predictive error rates using a prequential error estimator and the accuracy. We also implemented and computed the McNemar's Test to compare paired proportions of both algorithms classification results, and the statistic \emph{Q} proposed in \cite{Gama2013} to comparative assessment between any two algorithms. The latter statistic allows to compare the performance of two algorithms from the sequences of the prequential accumulated loss for each algorithm: $ {S_i^A } $ and $ {S_i^B } $,
\beq Q_i (A,B) = \log ({{S_i^A } \over {S_i^B }}) \eeq
and by using fading factors the $ Q_i  $ statistic takes the form: 
\beq Q_i^\alpha  (A,B) = \log ({{L_i (A) + \alpha  \times S_{i - 1}^A } \over {L_i (B) + \alpha  \times S_{i - 1}^B }}) \eeq
where $ L_i (A) $ and $ L_i (B) $ are the computed loss for the current instance and $ {\alpha  \times S_{i - 1}^A } $, $ {\alpha  \times S_{i - 1}^B } $ are the corresponding fading accumulated losses. The signal of $ Q_i $ is informative about the relative performance of both models, while its value shows the strength of the differences.

In order to assess our approach and the validity of the statements we made, we conducted a set of experiments on three synthetic and one real data sets. For each of the experiments performed on the synthetic data sets we computed and plotted the comparative metrics:  prequential accuracy, and \emph{Q} statistic. Additionally, in the case of the experiment on the real world problem data set, we compared paired proportions of both algorithms classification results with the McNemar's Test and we also computed and plotted the predictive accuracy and the percent error rate.
The first five data sets shown in Table \ref{table:Table1} are synthetic and  available in MOA. This framework facilitates the reproducibility of the experiments. In those cases the experiments were performed over a total of $100,000$ instances for each data set. The Fading Factor Classification Performance Evaluator was used with a fading factor ($\alpha $) value of $0.999$ for data streams without concept drift, while in the concept drifting data streams the fading factor was $0.95$. Both algorithms were evaluated with setups corresponding to its default values \cite{2012iblstreams}. The results regarding the metrics above indicated show that the performance of Online-KISSME-Stream in the data sets in which concept drift does not occur are better than those yielded by the IBLStream. This is evidenced by the comparative prequential accuracy where Online-KISSME-Stream shows better results than IBLStream, even from the early stages. Likewise, the same holds for the comparative predictive prequential error. In  this case, the sign of \emph{Q} is mainly negative, illustrating the overall advantage of our proposal over IBLStream. In the concept drifting data streams although the performance of Online-KISSME-Stream is not as good compared to IBLStream still the results yielded by our approach can be considered as competitive. The results of the evaluations are depicted  graphically from Figure~\ref{fig:one} to Figure~\ref{fig:four}.
We also evaluated the algorithms on KDD Cup 99. It is a well-known real world problem data set regarding network intrusion detection. In \cite{rivero2014} different preprocessing techniques that have been applied to this problem, prior to evaluate machine learning algorithms, have been reviewed. This data set is commonly used in the research community  because it is available, labeled and preprocessed. 
The original data set contains about $5$ million instances, each of which represents a TCP/IP connection made up of $41$ numeric and nominal attributes. In many investigations a small portion representing $10\%$ of the original data set is used, containing $494,021$ instances. In our experiments we used $111,000$ instances of the $10\%$ KDD Cup 99 data set.
\subsection{Evaluation Results on Synthetic Datasets}
\begin{figure*}
\centering
\scalebox{1.3}{\includegraphics{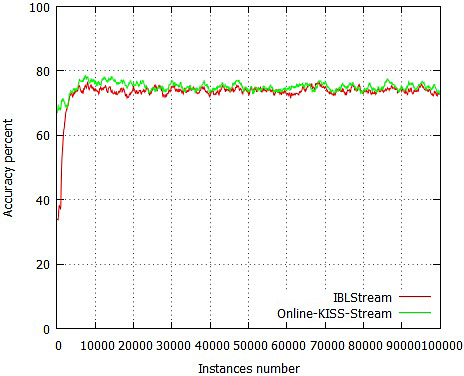}}
\hspace*{0cm}
\scalebox{1.1}{\includegraphics{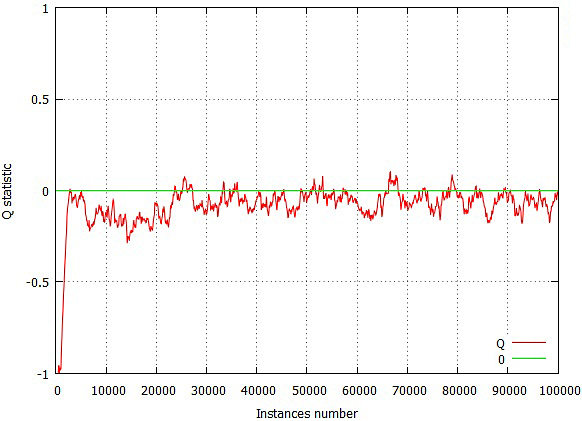}}
\caption{Accuracy and \emph{Q} statistic (Online-KISSME-Stream/IBLStream) for Random Tree Generator.}
\label{fig:one}
\end{figure*}
\begin{figure*}
\centering
\scalebox{1.3}{\includegraphics{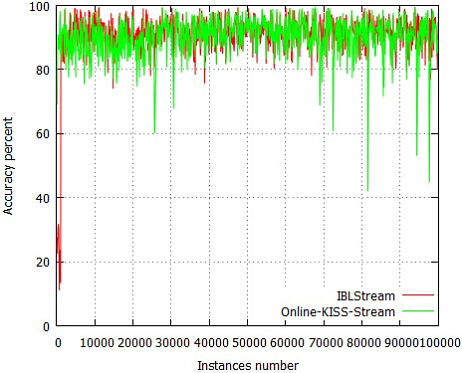}}
\hspace*{0cm}
\scalebox{1.1}{\includegraphics{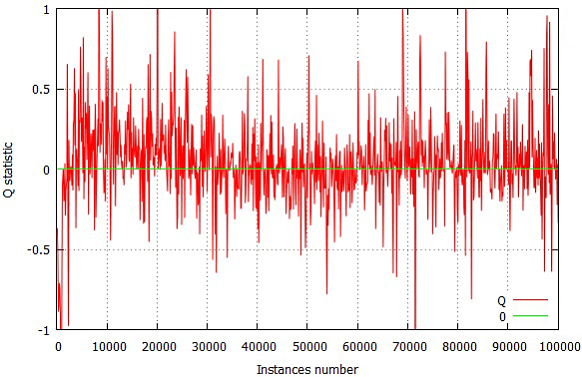}}
\caption{Prequential accuracy and \emph{Q} statistic (Online-KISSME-Stream/IBLStream) for Rotating Hyperplane.}
\label{fig:three}
\end{figure*}

On the Random Tree Generator with $100,000$ instances the prequential accuracy (see Figure~\ref{fig:one}) of Online-KISSME-Stream is higher than the IBLStream one. This leads to have a minor prequential error than IBLStream. The \emph{Q} statistic (see Figure~\ref{fig:one}) shows a higher area under the curve for the values less than zero meaning that Online-KISSME-Stream had less losses on this data set.
\begin{figure*}
\centering
\scalebox{1.4}{\includegraphics{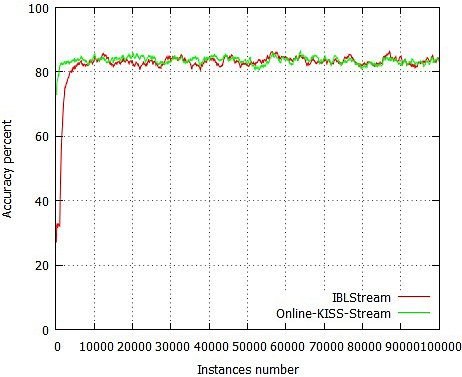}}
\hspace*{0cm}
\scalebox{1.2}{\includegraphics{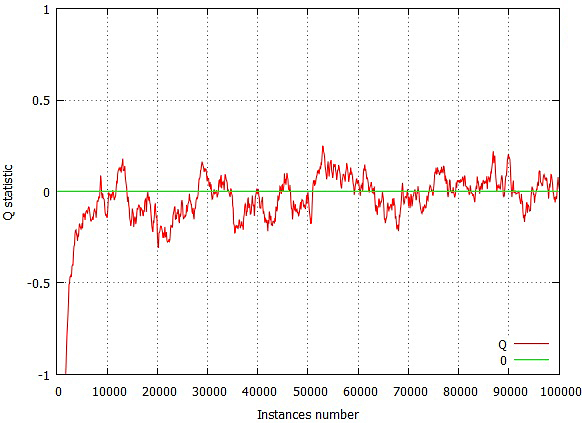}}
\caption{Prequential accuracy and \emph{Q} statistic (Online-KISSME-Stream/IBLStream) for Waveform.}
\label{fig:two}
\end{figure*}
In the Waveform data set's instances evaluated by Online-KISSME-Stream algorithm we observe that the prequential accuracy (see Figure~\ref{fig:two}) is higher than IBLStream one, specifically for the earlier instances. Likewise, the prequential error is less than the IBLStream one; and the area under the curve for values less than zero in the \emph{Q} statistic (see Figure~\ref{fig:two}) is greater, meaning less losses in this data set too.
On the Rotating Hyperplane data set the prequential accuracy results (see Figure~\ref{fig:three}), the prequential error  and the \emph{Q} statistic (see Figure~\ref{fig:three}) show very competitive results of classification, with a slight advantage for IBLStream classification algorithm. 

\subsection{Evaluation Results on KDD Cup 99 Dataset}
\label{sec:kdd}
\begin{figure*}
\centering
\scalebox{0.3}{\includegraphics{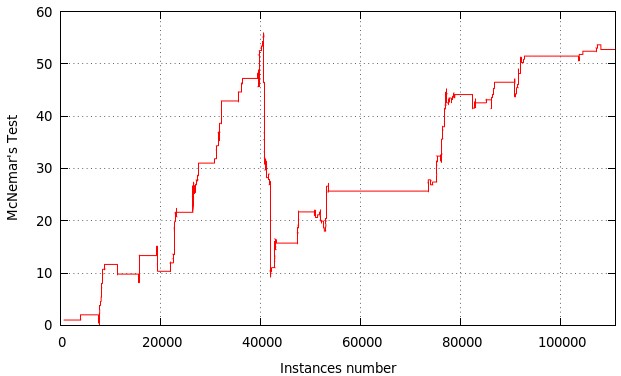}}
\hspace*{0cm}
\scalebox{1.3}{\includegraphics{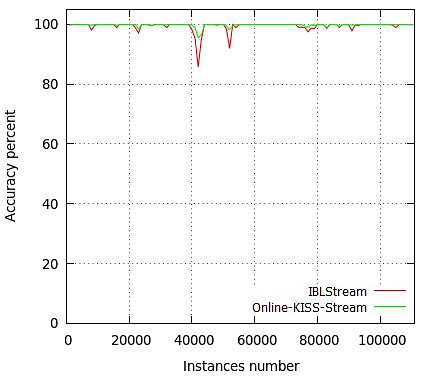}}
\caption{Prequential accuracy and \emph{Q} statistic (Online-KISSME-Stream/IBLStream) for KDD Cup 99.}
\label{fig:four}
\end{figure*}

Two experiments were performed on $111,000$ KDD Cup 99 instances. In the first, the prequential accuracy of Online-KISS-Stream was better than the IBLStream (Figure \ref{fig:four}). In the second, we applied the McNemar's Test. This  test is non-parametric on nominal data, which has been widely used for the comparison of batch learning classification algorithms. In \cite{Gama2013} the applicability of this test to classification problems in data stream environments  is emphasized. This test has acceptable type I error. To implement both quantities $n_{i,j}:n_{0,1}$ denotes the number of examples misclassified by Online-KISSME-Stream and that were not by IBLStream;  whereas  $n_{i,j}:n_{1,0}$ denotes the number of examples misclassified by IBLStream and that were not by Online-KISSME-Stream. Then, two hypotheses were defined: $H_0$ no differences between classifiers and $H_1$ there are differences between classifiers. The null hypothesis $H_0$ is rejected if with one degree of freedom and confidence level of $0.99$, the statistic is greater than $6.635$. Along with ca. the first $8000$ instances $H_0$ is accepted showing no differences in classification since the value of the McNemar statistic is less than $6.635$. This value becomes greater afterwards and $H_0$ is rejected indicating differences between both classifiers (see Figure~\ref{fig:four}). This test along with the prequential accuracy reinforces that our proposal is better in this real world problem.

\section{Conclusions}
\label{sec:conclusions}
The distance metric learning has attracted great interest from the research community in recent years. Most approaches define an optimization model combining the constraints of similarity with dissimilarity as loss functions with regularizers. There are five properties that cast these algorithms for certain environments. An important characteristic that led us to think about the development of this work is the scalability which is essential to ensure incremental learning in streaming scenarios. In this paper we proposed Online-KISS-Stream, a new instance-based data stream classification algorithm that learns a Mahalanobis metric based on KISSME algorithm. The online optimization proposed by KISSME computes the similarity and dissimilarity matrices from outer vector product and then computes the eigen decomposition of the difference of the inverse of both matrices, allowing to obtain the Mahalanobis matrix in an online way. Furthermore, by combining with the concept drift detection it updates the Mahalanobis matrix whenever required.
To evaluate the performance of Online-KISS-Stream several experiments were performed on synthetic and real world data sets. We compared successfully the results yielded by our approach with IBLStream. 
We implemented the established metrics for data stream classification algorithms such as  prequential error,  prequential accuracy, and \emph{Q} with fading factors as forgetting mechanism. For statistical significance we used the McNemar's Test in the KDD Cup 99 data set showing differences between our algorithm and IBLStream. Future work will address distance metric learning with different regularizers.
\section*{Acknowledgment} Erasmus Mundus Action 2 Programme, Eureka SD Project is gratefully acknowledged for funding.
%
\bibliographystyle{IEEEtran}
\bibliography{KISS-Stream}
\end{document}